# A New Algorithm for Hidden Markov Models Learning Problem

Taha Mansouri[1], Mohamadreza Sadeghimoghadam[2*], Iman Ghasemian Sahebi[3]

1. Postdoctoral Researcher, School of Science, Engineering, and Environment, University of Salford

2. Associate Professor of Industrial Management, Faculty of Management, University of Tehran

3. PhD Candidate of Operation Management, Faculty of Management, University of Tehran

**Abstract**

This research focuses on the algorithms and approaches for learning Hidden Markov Models (HMMs) and compares HMM learning methods and algorithms. HMM is a statistical Markov model in which the system being modeled is assumed to be a Markov process. One of the essential characteristics of HMMs is their learning capabilities. Learning algorithms are introduced to overcome this inconvenience. One of the main problems of the newly proposed algorithms is their validation. This research aims by using the theoretical and experimental analysis to 1) compare HMMs learning algorithms proposed in the literature, 2) provide a validation tool for new HMM learning algorithms, and 3) present a new algorithm called Asexual Reproduction Optimization (ARO) with one of its extensions – Modified ARO (MARO) – as a novel HMM learning algorithm to use the validation tool proposed. According to the literature findings, it seems that population-based algorithms perform better among HMMs learning approaches than other algorithms. Also, the testing was done in nine benchmark datasets. The results show that MARO outperforms different algorithms in objective functions in terms of accuracy and robustness.

# Contents

| List of Abbreviations | |
|---|---|
| HMMs | Hidden Markov Models |
| ARO | Asexual Reproduction Optimization |
| MARO | Modified Asexual Reproduction Optimization |
| BW | Baum-Welch Algorithm |
| SoP | Sum-of-Pairs |
| MLE | Maximizing Likelihood Estimation |
| MMD | Minimizing the Model Divergence |
| MPE | Minimizing the output or state Prediction Error |
| VAR | Vector Auto Regressive |
| MSA | Multiple Sequence Alignment |
| GA | Genetic Algorithm |
| VA | Viterbi Algorithm |
| EM | Expectation Maximization |
| LSEQ | Length of Sequences |
| LO | Log-Odd |

## 1. Introduction

Hidden Markov Models (HMMs) is a stochastic model for sequential data. It is a stochastic process determined by the two interrelated mechanisms: 1) a hidden markov chain with a limited number of states and a group of observation probability distributions, each associated with a state. At each discrete time instant, the process does assume to be in a condition, and observation does generate by the probability distribution corresponding to the current state [1]–[4]. HMMs are probabilistic models that capture statistical regularities within a linear sequence [5]. As we will see, HMMs are very well suited for the mentioned tasks. Indeed, HMMs have been largely used in many different problems in molecular biology, including gene finding, profile searches, multiple sequence alignment, identification of regulatory sites, or the prediction of secondary structure in proteins [6]. A general HMM is illustrated in Figure 1, where the exact form of interactions within states and within observables is abstracted.

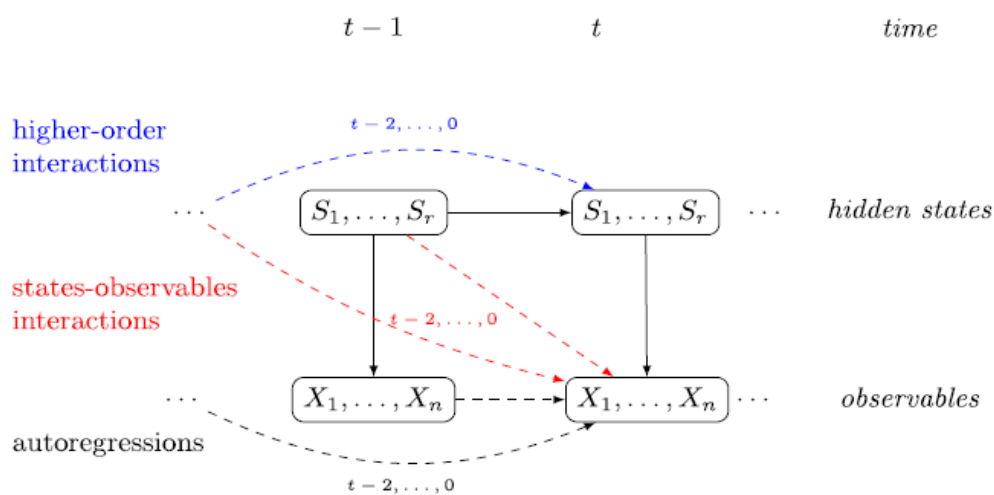

**Fig.1.** An abstracted general HMM [7]

Theoretical and empirical results have shown that, given an adequate number of states and a sufficiently rich set of data, HMMs can represent probability distributions corresponding to complex real-world phenomena in terms of simple and compact models [8], [9]. This is supported by the success of HMMs in various practical applications [10], where it has become a predominant methodology for design of automatic speech recognition systems [11], [12], signature verification [13], [14], communication and control [1], [15], bioinformatics [16], [17], computer vision [18], [19], computer and network security [20]–[22] and anomaly detection or in misuse detection [4], [20], [23].

In many practical applications, the collection and analysis of training data are expensive and time-consuming. Consequently, data for training an HMM is often limited in practice and may over time no longer represent the underlying data distribution. However, the performance of HMMs model depends on the availability of an adequate amount of representative training data to forecast its parameters, and in some cases, its topology. In static environments, where the underlying data distribution remains fixed, designing an HMM with a limited number of training observations may significantly degrade performance. An HMM trained using data sampled from the environment will incorporate some uncertainty concerning the underlying data distribution [24].

It is common to acquire additional training data from the environment at some point in time after a pattern classification system has been initially trained and deployed for operations. Since limited training data is typically employed in practice, and hidden data distribution is susceptible to change, a system based on HMMs should allow adaptation in response to new training data from the operational environment or other sources. The ability to efficiently adapt HMM parameters in response to newly-acquired training data through incremental learning is, therefore, an undisputed asset for sustaining a high level of performance. Indeed, refining an HMM to novelty encountered in the environment may reduce its uncertainty concerning the underlying data distribution.

All new proposed methodologies and algorithms in HMM learning is required to be validated. One way to validate learning algorithms is benchmarking their performance in a specific dataset with other algorithms in the same or different datasets. This study uses two kinds of comparisons as a validation tool for future researchers. The first comparison includes reviewing benchmark open datasets and objective functions in the literature and comparing their results. The best-performed algorithms in the literature are determined in this phase. The main contributions of this paper are three-fold. The first contribution is the comprehensive benchmark of algorithms and datasets in the literature. These techniques are classified according to the objective function, optimization technique, and target application. An analysis of their convergence properties and of their time and memory complexity was presented. Experiments show that MARO overcomes many problems encountered by other learning algorithms. As shown in the experiments, in the case of convergence speed (which is very low in hybrid algorithms), MARO is a high-speed algorithm. Also, MARO is very fast in terms of execution time; it obtained the shortest execution time. Being a single solution (unlike population-based algorithms) calls the objective function once, and thus the execution time is fast. In terms of exploit power, because it inherits from the previous generation and searches the last path, MARO outperforms population-based algorithms. MARO overcomes the problem of being trapped in the local optimum (typically in Hebbian-based algorithms) by following the simulated annealing algorithm rule and searching the broader search range. Finally, unlike many population-based algorithms, MARO tunes itself based on its path, and it does not need parameter setting. The applicability of these techniques is assessed for incremental learning scenarios in which new data is either abundant or limited. Finally, these techniques' advantages and shortcomings are outlined, providing the key issues and guidelines for their application in different learning scenarios.

The remainder of this paper is organized as follows. Section 2 delivers a brief introduction to HMMS. It includes definitions and historical backgrounds, formulation and construction process and outlines various learning approaches and algorithms introduced so far for HMMs. Section 3 presents a novel HMM learning algorithm accompanied by its extension is also introduced. Section 4 shows the experimental approach where the benchmark datasets from the literature, their objective functions, and theoretical and empirical findings. Finally, Section 5 concludes and suggests future study gaps.

## 2. Background and Literature Review
### 2.1. Element of HMMs

HMM is a stochastic process determined by the two interrelated mechanisms, an underlying Markov chain having a finite number of states and have a set of random functions. Each of these is associated with one condition. At each discrete time instant, the process is assumed to be in a state, and observation is generated by the random functions corresponding to the current state. According to the current state's transition probability matrix, the hidden Markov chain then transits to the next state. The following model parameters completely characterize the formal specifications of an HMM with continuous observation densities, and we used the same notation as Rabiner [25]:

1. $N$ is the number of states,

2. $M$ is the number of mixtures in the random function,

3. A is the transition probability distribution matrix

$A = \{a_{ij}\}$, where $a_{ij}$ is the transition probability of the Markov chain transiting to state $j$, given the current state $i$, that is:

$$a_{ij} = P[q_{t+1} = j | q_t = i], \tag{1}$$

Where $q_t$ is the state at time $t$ and $a_{ij}$ has the following properties:

$$\begin{aligned} &a_{ij} \geq 0, \quad 1 \leq i,j \leq N, \\ &\sum_{j=1}^{N} a_{ij} = 1, \quad 1 \leq i \leq N; \end{aligned} \tag{2}$$

4. B is the observation probability distribution matrix, $B = \{b_j(o)\}$ where $b_j(o)$ is the random function associated with state $j$. The most general representation of the random function is a finite mixture of Gaussian distributions of the form.

$$b_j(o) = \sum_{k=1}^{M} c_{jk} G(0, \mu_{jk}, U_{jk}), \quad 1 \leq j \leq N; \tag{3}$$

Where $o$ is the observation vector, $c_{jk}$ is the mixture coefficient for the $k$th mixture in state $j$, and $G$ is the Gaussian distribution with mean vector $\mu_{jk}$ and covariance matrix $U_{jk}$ for the $k$th mixture component in state $j$. The mixture coefficient $c_{jk}$ has the following properties to satisfy the stochastic constraints:

$$\begin{aligned} &c_{jk} \geq 0, \quad 1 \leq j \leq N, 1 \leq k \leq M, \\ &\sum_{k=1}^{M} c_{jk} = 1, \quad 1 \leq j \leq N \end{aligned} \tag{4}$$

5. $\pi$ is the initial state distribution matrix, $\pi = \{\pi_i\}$ in which $\pi_i = P[q_1 = i], \ 1 \leq i \leq N;$

It can be seen that the elements of an HMM are the model parameters of $N$, $M$, $A$, $B$, and $\boldsymbol{\pi}$. However, the values of $N$ and $M$ are implicitly existed in the dimension of the matrix $A$ and $B$, respectively. For convenience, we represent the HMM with the following notation $\lambda = (A, B, \pi)$.

### 2.2. HMM Learning Algorithm

Before using HMM for any problem, an HMM must be trained. Many criteria can be used for this problem. The most common one is to maximize the probability of the observation sequence, $O = (o_1 \; o_2 \; ... \; o_T)$, generated by the given HMM $\lambda$, i.e., $P[O|\lambda]$. The Baum-Welch Algorithm developed by Baum is one of the most successful optimization methods for this problem. The Baum-Welch Algorithm is a hill-climbing algorithm that includes a set of learning formulas and guarantees that the learning HMM $\bar{\lambda}$ will equal to or better than the initial HMM model, $\lambda$, i.e., $P[O|\bar{\lambda}] \geq P[O|\lambda]$. The re-estimation formulas are shown as follows:

$$\bar{\pi} = \alpha_1(i)\beta_1(i) / \sum_{j=1}^{N} \alpha_T(j), \tag{5}$$

$$\bar{a}_{ij} = \sum_{t=1}^{T} \alpha_{t-1}(i) a_{ij} b_j(o_t) \beta_t(j) / \sum_{t=1}^{T} \alpha_{t-1}(i) \beta_{t-1}(i), \tag{6}$$

$$\bar{c}_{jk} = \sum_{t=1}^{T} \gamma_t(j,k) / \sum_{j=1}^{T} \sum_{k=1}^{M} \gamma_t(j,k), \tag{7}$$

$$\bar{\mu}_{jk} = \sum_{t=1}^{T} \gamma_t(j,k) \cdot \boldsymbol{o}_t / \sum_{t=1}^{T} \gamma_t(j,k), \tag{8}$$

$$\bar{U}_{ij} = \sum_{i=1}^{T} \gamma_t(j,k) \cdot (\boldsymbol{o}_t - \mu_{jk})(\boldsymbol{o}_t - \mu_{jk})^t / \sum_{t=1}^{T} \gamma_t(j,k), \tag{9}$$

Where $\bar{\pi}$, $\bar{a}_{ij}$, $\bar{c}_{jk}$, $\bar{\mu}_{jk}$ and $\bar{U}_{ij}$ are the model parameters of $\bar{\lambda}$, $\gamma_t(j,k)$ is the probability of being in state $j$ at time $t$ with $k$th mixture component accounting for $o_t$ of the form:

$$\gamma_t(j,k) = \left[\frac{\alpha_t(j)\beta_t(j)}{\sum_{i=1}^{N}\alpha_t(i)\beta_t(i)}\right]\left[\frac{c_{jk}G(o_t,\mu_{jk},U_{jk})}{\sum_{m=1}^{M}c_{jm}G(o_t,\mu_{jm},U_{jm})}\right], \quad (10)$$

And $\alpha_t(i)$ is the forward variable [3] of the form:

$$\alpha_t(i) = \begin{cases} \pi_i b_i(o_i), & t=1 \quad 1 \leq i \leq N, \\ \left[\sum_{i=1}^{N}\alpha_{t-1}(i)a_{ij}\right]b_j(o_t), & 1 < t \leq T, 1 \leq i \leq N \end{cases} \quad (11)$$

And $\beta_t(i)$ is the backward variable [3] of the form:

$$\beta_t(i) = \begin{cases} 1, & t=T, 1 \leq i \leq N, \\ \sum_{j=1}^{N}a_{ij}b_j(o_{t+1})\beta_{t+1}(j), & 1 \leq t \leq T, 1 \leq i \leq N \end{cases} \quad (12)$$

With the Baum-Welch (BW) Algorithm, we can improve the model parameters by the repetition of the substitution of $\lambda = \bar{\lambda}$. It is a problem with the BW Algorithm being a hill-climbing algorithm, and any arbitrary chosen initial model j will usually lead to a sub-optimal model. In the following we will use the term *test sequence* when we refer to a sequence obtained by a dataset, and *reference sequence* for a manually refined sequence that is believed to be of high quality. We used two different methods for scoring the test sequence obtained during the experiments. The first method does not rely on any form of prior knowledge regarding the structure of the resulting sequence, whereas the second method requires a reference sequence for comparison. Both scoring methods are based on the widely used sum-of-pairs scoring function. In experiments where no prior knowledge of the structure of the resulting dataset was available, we used the standard sum-of-pairs (SoP) score shown in Eq. (13):

$$SoP = \sum_{i=1}^{n-1}\sum_{j=i+1}^{n}D(l_i,l_j) \quad (13)$$

Where $l_i$ is sequence $i$ and $D$ is a distance metric. During the training of the HMM its quality needs to be measured. For this a log-odds score is used, which is based on a log-likelihood score [3] shown in Eq. (14):

$$Log-odds\,(O,\lambda) = \frac{1}{N}\sum_{i=1}^{N}\log_2\frac{P(O_i|\lambda)}{P(O_i|\lambda_N)} \tag{14}$$

Where $O = \{O_1, O_2, \ldots, O_N\}$, is the set of sequences, $\lambda$ is the trained HMM, and $\lambda_N$ is a null-hypothesis model.

### 2.3. Analyzing the HMM Learning Algorithms

Several algorithms from the literature were applied to the learning of HMM parameters. Figure 2 presents a taxonomy of techniques for learning HMM parameters according to the objective function, optimization technique, and target application. As shown, they fall in the categories of standard numerical optimization, expectation-maximization, and recursive estimation, with the objective of either maximizing likelihood estimation (MLE) criterion, minimizing the model divergence (MMD) of parameters penalized with the MLE, or minimizing the output or state prediction error (MPE). The target application implies a scenario for data organization and learning. Some techniques have been designed for block-wise estimation of HMM parameters, while others for symbol-wise estimation of parameters. Block-wise techniques are designed for scenarios in which training symbols are organized into blocks of sub-sequences. The HMM re-estimates its parameters after observing each sub-sequence. In contrast, symbol-wise techniques, also known as recursive or sequential techniques, are designed for scenarios in which training symbols are observed one at a time, from a stream of characters. The HMM parameters are re-estimated upon observing each new character. The rest of this section provides a survey of techniques for line learning of HMM parameters shown in Fig. 2.

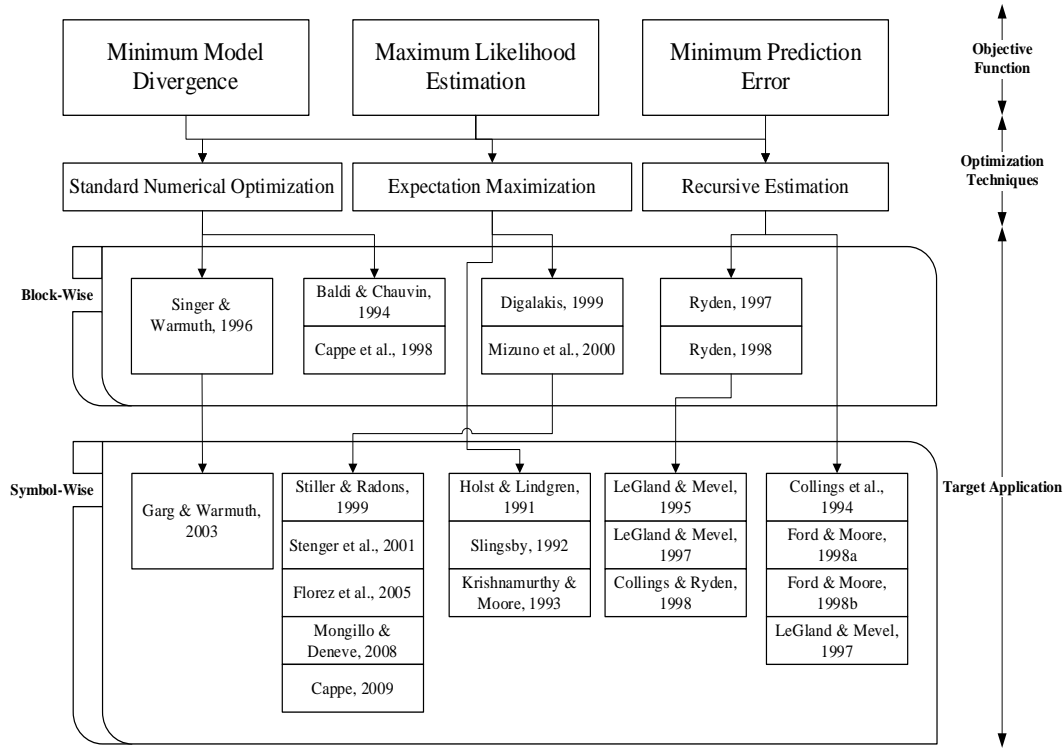

**Fig. 2**. Taxonomy of techniques for the learning of HMM parameters, adapted from [4]

While good models and learning algorithms play a crucial role in time series analysis, a significant challenge in quite a few scientific dynamic modeling tasks, as pointed out by [26], turns out to be collecting reliable time-series data. In some situations, the dynamic process of interest may evolve slowly over time, such as Alzheimer's disease progression. It may take months or even years to obtain enough time series data for analysis. In other situations, it may be very difficult to measure the dynamic process of interest repetitively due to the measurement technique's destructive nature. One such example is the gene expression time series. Although obtaining reliable time series, or dynamic data, can be difficult, it is often easier to collect static, order fewer snapshots of the dynamic process of interest. For example, doctors can collect many samples from a current pool of Alzheimer's patients in possibly different disease stages. Scientists can quickly obtain large amounts of static gene expression data from multiple experiments. Huang and Schneider [26] propose an estimator for the vector autoregressive (VAR) model that uses dynamic and static data and derives a simple gradient-descent algorithm from minimizing its non-convex estimation objective. Through simulations and experiments on video data, they demonstrate that static data does improve estimation, especially when the amount of dynamic data is small.

The subspace identification methods used in control theory use spectral approaches to discover the relationship between hidden states and the observations. In this literature, the association is discovered for linear dynamical systems such as Kalman filters [27]. The basic idea is that the relationship between observations and hidden states can often be found by spectral methods correlating the past and future observations. However, algorithms used in the literature cannot be directly utilized to learn HMMs because they assume additive noise models with noise distributions independent of the underlying states. Such models are not suitable for HMMs (an exception is [28].

The second idea is that we can represent the probability of sequences as products of matrix operators, as in the literature on multiplicity automata [29], [30]. This idea was re-used in both the Observable Operator Model of [31] and the Predictive State Representations of [32] related and both of which can model HMMs. The former work by [31] provides a non-iterative algorithm for learning HMMs, with asymptotic analysis. However, this algorithm assumed knowing a set of 'characteristic events,' which is a relatively strong assumption that effectively reveals some relationship between the hidden states and observations. In our algorithm, this problem is avoided through the first idea. Some of the work techniques in [33] for tracking belief states in an HMM are used here. As discussed earlier, we provide a result showing how the model's conditional distributions over observations (conditioned on a history) do not asymptotically diverge. This result was proven in [33] when an approximate model is already known. Roughly speaking, the reason this error does not diverge is that the previous observations are always revealing information about the next state; so, with some appropriate contraction property, we would not expect our errors to diverge. Among recent efforts in various communities [28], [34], [35], the only previous efficient algorithm shown to learn HMMs in a setting similar to ours is due to [36]. Their Algorithm for HMMs is a specialization of a more general method for learning phylogenetic trees from leaf observations. While both this algorithm and ours rely on the same rank condition and compute similar statistics, they differ in two significant regards. First, [36] were not concerned with large observation spaces, and thus their algorithm assumes the state and observation spaces to have the same dimension. Also, [36] take the more ambitious approach of learning the observation and transition matrices explicitly, which unfortunately results in a less stable and less sample-efficient algorithm that injects noise to spread apart the spectrum of a probability matrix artificially. In table 1 illustrated the important related works.

**Table1**

HMM learning algorithms

| Author (s) | Problem | Algorithm | Dataset |
|---|---|---|---|
| [7] | (Multiple sequence alignment) alignment of protein sequences. | hybrid algorithm combining PSO with evolutionary algorithms | Pfam database BAliBASE database |
| [37] | Genome sequencing (DNA sequencing) | EM algorithm | - |
| [38] | Epileptic patients' prediction | SA-EM algorithm | sequences of daily seizures in a sample of 507 epileptic patients |
| [39] | natural language processing | Spectral algorithm | - |
| [40] | speech recognition | GA and Baum-Welch algorithms | TMIIT database |
| [41] | Hand gesture recognition | Baum-Welch algorithm Artificial Bee Colony (ABC) algorithm | Cambridge's hand gesture data |
| [42] | inventory management | modified Baum–Welch algorithm with limiting distribution (BWLC algorithm) | Numerical Example |
| [43] | Computer Memory Decoding | Forward Filtering Backward Smoothing (EFFBS) algorithm | Numerical Example |
| [44] | polyphonic music | Recurrent Neural Network-Restricted Boltzmann Machine (RNN-RBM) | polyphonic music datasets |
| [45] | predictive maintenance | Baum-Welch/ Segmental K-means learning Algorithm | simulated industrial process |
| [46] | system identification | Spectral Algorithm | Simulated data |
| [47] | - | Modified Baum-Welch | simulated data |
| [48] | multiple sequence alignment | random drift particle swarm optimization (RDPSO) algorithm | nucleotide data set amino-acid data set |
| [49] | handwriting recognition | Modified Baum-Welch Algorithm | open-source repository |
| [50] | Speech Recognition | Chaos Optimization algorithm with BW | FARSDAT dataset |
| [51] | Speech Recognition | deep belief network pre-training algorithm | challenging business search dataset |
| [52] | Speech Recognition | genetic algorithm | Data generated with word recognizer |
| [53] | Alphabet Recognition | Alpha-Em Algorithm | simulated data |
| [54] | - | Randomized Search Algorithms Metropolis Algorithm | Simulated data |
| [55] | Automatic speech recognition | GA-BW | Mixt-Gaussian speech database |
| [56] | - | Baum-Welch algorithm with incomplete data | Simulated data |
| [57] | biological sequence modeling | BW-PSO | lung cancer-related genes dataset |

| Author (s) | Problem | Algorithm | Dataset |
|---|---|---|---|
| [58] | speech recognition | PSO-GA | Census (AN4) database |
| [59] | speech recognition | GA-BW | Simulated data |
| [60] | speech recognition | ANN | TIMIT continuous speech data |
| [61] | biological and inflation | BW-V | Pfam database Inflation database (from 2001 to 2015) in Iran |

Following definitions by [62] and [63], an HMM that performs incremental learning can independently learn one new block of training data at a time. With incremental learning, HMM parameters should be efficiently updated from recent data without requiring access to the previously-learned training data. Besides, parameters should be corrected without corrupting previously-acquired knowledge. Standard techniques for estimating HMM parameters involve batch learning, based either on specialized Expectation-Maximization (EM) methods [64], such as the BW Algorithm [65] or on numerical optimization techniques, such as the Gradient Descent algorithm [66]. In either case, HMM parameters are estimated over several training iterations until some objective function, e.g., maximum likelihood over some independent validation data, is maximized. For a batch learning technique, a fixed-length sequence $O = o_1, o_2…, o_T$ of T training observations $o_i$ is assumed to be available throughout the training process. Taking that O is assembled into a block D of training data, each training iteration typically involves observing all sub-sequences in D before updating HMM parameters.

As an alternative, several on-line learning techniques proposed in literature may be applied for learning. These include numerical optimization and recursive estimation methods and assume the observation of a stream of data. Some of these techniques are designed to update HMM parameters at a symbol level (symbol-wise), while others update parameters at a sequence level (block-wise). Methods for on-line symbol-wise learning, also referred to as recursive or sequential estimation techniques, are designed for situations in which training symbols are received one at a time, and HMM parameters are re-estimated upon observing each new symbol. Methods for on-line block-wise learning are designed for situations in which training symbols are organized into a block of one or more sub-sequences. HMM parameters are re-estimated upon observing each new sub-sequence of characters. In either case, HMM parameters are updated from new training data without requiring access to the previously-learned training data and potentially without corrupting previously acquired knowledge. Table 2 shows a list of learning algorithms and their category in the literature so far.

**Table2**

HMM learning algorithm category

| Category | Learning Algorithm | Author(s) |
|---|---|---|
| Expectation–Maximization | Baum-Welch (BW) | [42], [50] |
| | Gradient Descent algorithm | [67] |
| | Modified Baum-Welch (MBW) | [47] |
| | Alpha-EM Algorithm | [53]a |
| Population-based | Particle Swarm Optimization (PSO) | [57] |
| | Genetic Algorithm (GA) | [40][52] |
| | Randomized Search Algorithms | [54] |
| | Metropolis Algorithm | [68] |
| | Memetic Particle Swarm Optimization (MPSO) | [69] |
| | Simulated Annealing (SA) | [70]–[72] |
| | Chaotic Simulated Annealing (CSA) | [50] |
| | Ants Colony Optimization (ACO) | [73] |
| | Artificial Bee Colony (ABC) | [41] |
| Hybrid | BW-PSO | [58] |
| | PSO-GA | [57][74] |
| | GA-BW | [59][55] |
| Others | ANN | [65][44] |
| | Robbins–Monro | [75], [76] |

The following table also shows the results of the superiority of learning algorithms in different datasets.

**Table 3**

Comparison of learning algorithms results in different datasets

| Author(s) | Algorithm | Dataset | Results | OF |
|---|---|---|---|---|
| [77] | PSO, SA | Pfam/ Balibase | PSO>BW>SA | Log- odds |
| [59] | GA | TMIIT | GA>BW | Log- odds |
| [41] | ABC | Cambridge hand gesture data | ABC>BW | SoP |
| [44] | B-ANN | Polyphonic music | B-ANN> G-ANN | SoP |
| [48] | RD-PSO | Amino-acid Dataset | RD-PSO>PSO> BW | Log- odds |
| [50] | PSO-BW | FARSDAT | PSO-BW>BW | Log- odds |
| [52] | GA | Synthetic | GA> BW | Log- odds |
| [54] | MA, SA | Synthetic | MA>SA> BW | SoP |
| [78] | PSO | Lung Cancer gene | PSO>BW | Log- odds |
| [58] | BW-GA, BW-PSO | Census (AN4) dataset | BW-GA> BW> BW-PSO | Log- odds |
| [60] | ANN | TMIIT | ANN> BF | Log- odds |

Note: OF: objective Function,

Table 4 shown the best algorithm results in HMMs datasets.

**Table 4**

Best algorithm results in HMMs datasets

| Dataset | Best Algorithm | Algorithm Type |
| --- | --- | --- |
| Pfam | SA-BW | Population-Based |
| TMIIT | SA | Population-Based |
| Cambridge hand gesture data | ABC | Population-Based |
| Amino-acid Dataset | RD-SA | Population-Based |
| Mixt-Gussian Speech | SA-BW | Population-Based |
| Census (AN4) dataset | GA-BW | Population-Based |
| Lung Cancer gene | SA | Population-Based |

### 2.4. Asexual Reproduction Optimization (ARO)

In addition to the learning algorithms existed in the literature, a novel algorithm with its extension for fine-tuning HMMs is proposed in this section. The algorithm is called ARO, and its extension is called Modified ARO (MARO). ARO is one of the optimization algorithms that originated from budding methods of asexual reproduction, recognized as a remarkable biological phenomenon [79], [80]. In this algorithm, every individual is represented by a binary string, as binary representation in evolutionary techniques.

A decision variable vector defines an ARO individual, $X = \{x_1, x_2, ..., x_n\} \mid X \in R^n$, and each variable is thought-out as the chromosome made by a set of bits that are called genes. Thus, a chromosome with a length of $L$ is considered that the first bit indicates the sign of the individual. The next $L_1$ bits represent the integer section whilst the last $L_2$ bits represent the decimal section of the chromosome. Therefore, $L = L_1 + L_2 + 1$ and the length of an individual becomes $L_1 = n \times L$. Fig. 3 illustrates the ARO chromosome.

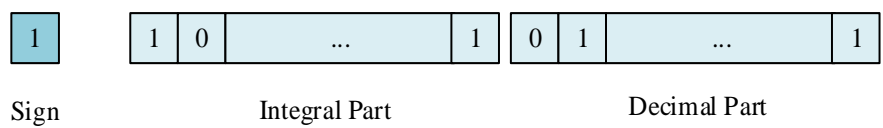

Sign    Integral Part    Decimal Part

**Fig 3**. An ARO chromosome

The algorithm will repeat as it is shown in pseudocode until the stopping criteria are satisfied. The stopping criteria for this algorithm are the number of iterations. For reproduction purposes, a copy of the parent called larva is produced. Then a substring with $g$ bits ($g \sim U[1, L_1]$) in the larva is randomly selected, where $U[1, L_1]$ is the uniform probability distribution between 1 and $L_1$. Then, the substring mutates bits so that in any selected gene, the number 1 is replaced by 0 and vice-versa. Afterward, for each bit of substring, randomly chosen from larva, a random number distributed ranged in [0, 1] is generated. If the generated number is less than 0.5, then the bit will be chosen from the parents; otherwise, it will be selected from the larva until the bud is completed. This process is illustrated in Fig. 4.

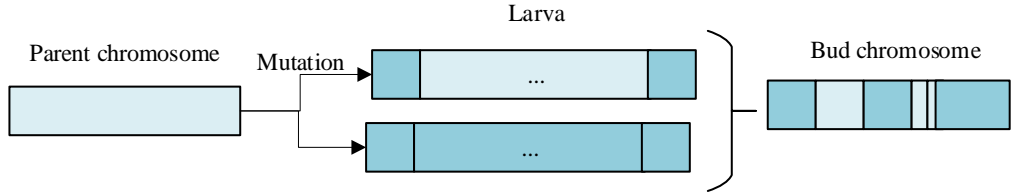

**Fig. 4**. Reproduction mechanism generated bud chromosome.

The probability of mutation is calculated as follows:

$$P = \frac{1}{\ln(g)} \tag{15}$$

Ultimately, the bud fitness is evaluated and compared with its parent fitness. At last, the most deserving is capable of subsisting to reproduce [81] was shown in Algorithm 1.

| **Algorithm 1**: Pseudocode of ARO |   |
|---|---|
| 1 | t = 1 // Initial time setting |
| 2 | P = initialize(L, U) // Parent initialization between lower and upper bound |
| 3 | Fitness P = Fitness(P) // Fitness of P is Calculated |
| 4 | while stopping conditions ≠ true do |
| 5 | Bud$_t$ = Reproduce (P); // P reproduce a Bud |
| 6 | Fitness Budt = Fitness Budt ; // Fitness of Bud (t) is calculated |
| 7 | if Fitness Budt > Fitness P then P = Bud(t); // Bud (t) is replaced by P |
| 8 | else Clear (Bud (t)); // Bud (t) is discarded |
| 9 | t = t + 1 |

Since ARO is a model-free algorithm, it can be applied for HMM learning. The weight matrix is converted to a vector and is randomly valued for the initialization phase. A block from the generated matrix is mutated based on ARO and is combined with the parent to produce a new weight. In the last version of ARO, some modifications have been applied. In this case, that bud is better than the parent. As a result, the bud is chosen, and the parent is discarded. But in Modified ARO (MARO), the bud error is tolerated. This tolerance should accomplish the following rules:

1) The degree of tolerance is high in the early stages of the algorithm, and low during the algorithm is ruining.
2) When the frequency of being trapped in the optimum local increases, the error tolerance increases and vice versa.

### 2.5. Suggested Algorithm

In ARO, the bud and the parent would be replaced only if the bud can improve (even an epsilon) the objective function. On the other side, our new proposed algorithm called MARO [79], [80]. Therefore, the following is added by MARO to ARO as shown in Algorithm 2.

**Algorithm 2:** Added by MARO to ARO

| 1 | if lb < lp + Δt then P = Bud(t); // Accept Bud(t) |

Where lb is the first error generated from the bud and lp is the amount of that error in the parent. $\Delta_t$ is computed as follows:

$$\Delta_t = \frac{\ln(loc)}{\sqrt{t}} \qquad (16)$$

Ln is a natural logarithm, loc the number of being trapped in the local optimum and t the number of iterations. The nature of this function is the acceptance of the number of local optimums with low steep. When the number of optimum local increases, the acceptance range will increase with lower steep. If it is equal to one, the minimum number of being trapped in the local optimum is zero; thus, we act the same as ARO. Iteration is the number of runs in the algorithm, and to reduce its effect in high iterations, we obtain its square root. We put the received number in the denominator to search narrower range when the iterations increase. The pseudocode of MARO is detailed in Algorithm 3.

**Algorithm 3:** Pseudocode of MARO

| 1 | t = 1 // Initial time setting |
| 2 | loc = 1 // Initial local number setting |
| 3 | = initialize(L, U) // Parent initialization between lower and upper bound |

| 4 | FitnessP = Fitness(P) // Fitness of P is Calculated |
| --- | --- |
| 5 | while stopping conditions ≠ true do |
| 6 | $Bud_t$ = Reproduce (P); // P reproduce a Bud |
| 7 | Fitness Budt = Fitness Budt ; // Fitness of Bud (t) is calculated |
| 8 | if Fitness Budt > Fitness P then P = Bud(t); // Bud (t) is replaced by P // Reset number of location |
| 9 | else if FitnessBudt > FitnessP-Δt then P = Bud(t); // Bud (t) is replaced by P // Δt is calculated via equation 4 |
| 10 | else Clear (Bud (t)); // Bud (t) is discarded |
| 11 | loc = loc + 1 |

Hence, ARO and MARO are both applied for HMM learning, and their results are compared to other algorithms in this study.

## 3. Experimental Set-up
### 3.1. Benchmark Datasets A

In this section, the most common open datasets used in the literature to test the algorithms are presented. The first dataset A consisted of three large sets of unaligned sequences for the three protein families Globin, Insulin, and Cytochrome C, which we extracted from the *Pfam* database [82] (see Table 5). These protein families have previously been used [83] and GA [84]. In our experiments, we used these data to compare our MARO versus BW and Simulated Annealing for HMM training. The score was used to judge the quality of the resulting alignments. The three protein family sets were divided into training and validation sets to check the HMM model's over-fitting. The size of each training set was 150, as shown in Table 5. The validation sets were the original datasets, excluding the training sets.

**Table 5**

The three protein families selected for dataset A

| Family | N | LSEQ (min, max) | T |
| --- | --- | --- | --- |
| Cytochrome C | 491 | 90 (19, 119) | 150 |
| Globin | 630 | 145 (121, 162) | 150 |
| Insulin | 248 | 75 (16, 168) | 100 |

*N*: number of sequences, LSEQ: length of sequences, and *T*: the size of the training set.

### 3.2. Benchmark Dataset B

While dataset A mostly served as an indicator of our algorithm's potential to align large sets of sequences compared with other HMM approaches, we were also interested in comparing the quality of the alignments with other non-HMM methods. For this purpose, we used data from the BAliBASE database [85], which contains several manually refined multiple sequence alignments designed explicitly for the evaluation and comparison of multiple sequence alignment methods. Even though the number of sequences set is relatively small and HMMs often require larger training sets, the framework we have based our work on has previously been compared to other methods using this benchmark database [86]. Due to the small number of sequences in each set, the sets were not split into training and validation sets. We selected a total of nine sequence sets from the first reference set from the BAliBASE database, which is listed in Table 6. Since reference alignments for these datasets exist, the sum-of-pairs score was used for determining the quality of the resulting test alignments.

**Table 6**

The nine benchmark datasets from BAliBASE

| Name   | N | LSEQ (min, max) | Identity (%) |
|--------|---|-----------------|--------------|
| 1idy   | 5 | (49, 58)        | <25          |
| 451c   | 5 | (70, 87)        | 20-40        |
| 1krn   | 5 | (66, 82)        | >35          |
| kinase | 5 | (263, 276)      | <25          |
| 1pii   | 4 | (247, 259)      | 20-40        |
| 5ptp   | 5 | (222, 245)      | >35          |
| 1ajsA  | 4 | (358, 387)      | <25          |
| glg    | 5 | (438, 486)      | 20-40        |
| 1taq   | 5 | (806, 928)      | >35          |

$N$: number of sequences, LSEQ: length of sequences.

For each set of sequences in datasets A and B, we conducted HMM training experiments with the Baum–Welch (BW), the Simulated Annealing (SA), the ARO with the log-odds score (LO-ARO), and the MARO with a sum-of-pairs score (SoP-MARO) as the objective function. Except for the BW, which is deterministic given a fixed initial HMM, we repeated all experiments 25 times for each training algorithm.

# 4. Results
## 4.1. Performance Comparison of HMM Learning Methods

Tables 6–9 summarize the average best fitness and standard error results for the HMM training and validation sets of dataset A. Tables 6 and 7 show the experiments' results. The log-odds score was used as the quality measure for the HMMs. The LO-ARO produced HMMs with better average log-odds scores than HMMs trained with BW and SA for all three training sets, whereas the SoP-MARO results were only comparable to BW-SA (Table 6).

The corresponding results achieved by applying the HMMs trained on the training sets to the related validation sets are shown in Table 6. Again, the LO-ARO achieved the best average scores, whereas the SoP-MARO scores were only comparable to the BW and SA scores.

Tables 7 and 8 summarize the results for the average best sum-of-pairs score for the alignments produced by the HMMs for the training and validation sets, respectively. Even though the LO-ARO achieved the best log-odds scores, the sum-of-pairs scores are only about as good as the scores for BW and SA (Table 6). In these experiments, the SoP-MARO achieved better scores than all other methods.

Regarding the validation sets, the LO-ARO seemed to yield only slightly better alignments than both BW and SA (Table 8). Again, the SoP-MARO was able to achieve better scores than all the other methods.

**Table 6**

Results for the training sets of dataset A

| Family | BW | SA | LO-ARO | SoP-MARO |
|---|---|---|---|---|
| Cytochrome C | 81.71 | 81.15±0.99 | 90.98±1.28 | 82.10±2.03 |
| Globin | 144.11 | 144.93±0.93 | 152.72±0.79 | 143.28±1.56 |
| Insulin | 132.15 | 128.63±1..07 | 139.85±0.63 | 130.50±1.05 |

*HMM log-odds scores ± standard error.*

**Table 7**

Results for the validation sets of dataset A

| Family | BW | SA | LO-ARO | SoP-MARO |
|---|---|---|---|---|
| Cytochrome C | 56.64 | 54.16±1.11 | 63.68±1.42 | 56.52±2.56 |
| Globin | 223.83 | 224.9±0.54 | 230.96±0.39 | 222.89±2.01 |
| Insulin | 87.67 | 84.37±0.98 | 93.60±1.02 | 88.20±2.92 |

*HMM log-odds scores ± standard error.*

**Table 8**

Sum-of-Pairs scores (divided by 1000) for the final alignments of the training sets of dataset A

| Family | BW | SA | LO-ARO | SoP-MARO |
|---|---|---|---|---|
| Cytochrome C | -197 | -202±1.5 | -195±1.2 | -190±0.9 |
| Globin | 45253 | 55952±103.4 | 45042±91.7 | 46013±90.3 |
| Insulin | 187 | 170±2.0 | 196±2.2 | 200±1.9 |

**Table 9**

Sum-of-Pairs scores (divided by 1000) for the final alignments of the validation sets of dataset A

| Family | BW | SA | LO-ARO | SoP-MARO |
|---|---|---|---|---|
| Cytochrome C | -853 | -863±2.1 | -795±2.0 | -783±1.4 |
| Globin | 41730 | 37656±117.8 | 41708±130.0 | 42.27±87.7 |
| Insulin | -827 | -826±2.3 | -817±2.1 | -788±2.0 |

As in the previous case, the SoP-MARO initially made an enormous improvement. Moreover, the SoP-MARO quickly achieved a better sum-of-pairs score than the LO-ARO and the other approaches.

### 4.2. Performance Comparison with Other Techniques

Table 9 shows the average best SoP scores obtained from the experiments performed on dataset B. The SoP-MARO results were better than those of the LO-ARO.

**Table 10**

Sum-of-Pairs (SoP) scores for the BAliBASE test sets

| Test Set | BW | SA | LO-ARO | SoP-MARO |
|---|---|---|---|---|
| 1idy | 0.516 | 0.139 | 0.516 | 0.624 |
| 451c | 0.264 | 0.347 | 0.338 | 0.650 |
| 1krn | 1.000 | 0.945 | 1.000 | 1.000 |
| kinase | 0.225 | 0.278 | 0.402 | 0.424 |
| 1pii | 0.165 | 0.193 | 0.268 | 0.662 |
| 5ptp | 0.634 | 0.743 | 0.756 | 0.885 |
| 1ajsA | 0.258 | 0.106 | 0.315 | 0.492 |
| glg | 0.740 | 0.690 | 0.769 | 0.888 |
| 1taq | 0.746 | 0.692 | 0.786 | 0.875 |

*N*: number of sequences, LSEQ: length of sequences.

Our algorithm combines the power of a fast, but not optimal, conventional heuristic (BW) with a stochastic heuristic (SoP-MARO and LO-ARO) that can improve suboptimal solutions further. Our comparison shows that the LO-ARO produced better scores than BW and SA for all test sets. This is quite remarkable since BW and SA are the most commonly used techniques for HMM training.

## 5. Discussion and Conclusion

This study provided a validation tool for researchers working on HMM learning algorithms. For this purpose, two analyses were employed. The first analysis reviewed the literature and used theoretical findings to compare different learning algorithms proposed in the literature. The second analysis was used for validating the theoretical analysis. Different algorithms from the literature were selected (based on the literature's performance) and experimented with some famous and open datasets. It was shown that population-based algorithms outperform other learning approaches according to the literature and experiments. The second part of this paper proposes a novel algorithm with one of its extensions for learning HMM models. This algorithm was also added to the comparison. The extension of the new proposed algorithm in this research (MARO) outperformed all algorithms.

From these experiments, we can conclude that the ARO with log-odds as the objective function (LO-MARO) increased both the log-odds score of the HMMs and the quality of the resulting alignments (with regards to a sum-of-pairs score) compared to the HMMs trained by BW and SA. The HMMs trained by the modified ARO with sum-of-pairs as an objective function (SoP-MARO) only had average log-odds scores comparable to those of BW and SA but produced better alignments than all of the other methods. [83] observed that there appeared to be a slight non-correlation between high log-odds scoring HMMs and good alignments in some of his test cases. Our experiments seem to verify this observation, assuming that the sum-of-pairs score is a useful indication of alignments' quality. It appears that further research on the topic of evaluating the quality of HMMs is needed. The ARO computation time is substantial compared to the BW and SA methods (4 h compared to a few minutes for large datasets and 2000 iterations). The vast majority of the computation time is spent on evaluating HMMs by the *forward*-function [3], which is done for each of the 20 particles in each iteration of the ARO and MARO. Since the learning of HMMs is usually done off-line, the proposed algorithm is still a viable alternative to the commonly used methods if higher scoring HMMs are needed. Since the MARO runtime is directly proportional to the number of chromosomes in the population, studies on the effect of smaller population sizes would be relevant.

Experiments show that MARO overcomes many problems encountered by other learning algorithms. As shown in the experiments, in the case of convergence speed (which is very low in hybrid algorithms), MARO is a high-speed algorithm. Also, MARO is very fast in terms of execution time, and in all datasets, it obtained the shortest execution time. Being a single solution (unlike population-based algorithms) calls the objective function once, and thus the execution time is fast. In terms of exploit power, because it inherits from the previous generation and searches the last path, MARO outperforms population-based algorithms. MARO overcomes the problem of being trapped in the local optimum by following the Simulated Annealing algorithm rule and searching the broader search range. Finally, unlike many population-based algorithms, MARO tunes itself based on its path, and it does not need parameter setting.

Finally, we would like to point out that the proposed algorithm is not limited to HMM training for sample datasets but serves as a general-purpose approach for optimizing HMMs. The described results show an excellent potential for other applications utilizing HMMs, such as genomic database searching [87] and speech recognition [3], [40].

**Declaration of competing interest**
The authors declare that they have no known competing financial interests or personal relationships that could have appeared to influence the work reported in this research.